%% file: anime_style_gan.tex
\documentclass{article}

\PassOptionsToPackage{numbers}{natbib}
\usepackage[preprint]{nips_2018}

\usepackage[utf8]{inputenc} 
\usepackage[T1]{fontenc}    
\usepackage{hyperref}       
\usepackage{url}            
\usepackage{booktabs}       
\usepackage{amsfonts}       
\usepackage{nicefrac}       
\usepackage{microtype}      

\usepackage{amsmath}
\usepackage[pdftex]{graphicx}

\title{Anime Style Space Exploration Using Metric Learning and Generative Adversarial Networks}

\author{
  Sitao Xiang\textsuperscript{1}\hspace{4em}Hao Li\textsuperscript{1, 2, 3}\\
  \textsuperscript{1}University of Southern California\\
  \textsuperscript{2}Pinscreen\\
  \textsuperscript{3}USC Institute for Creative Technologies\\
  \texttt{sitaoxia@usc.edu\hspace{2em}hao@hao-li.com} \\
}

\begin{document}

\maketitle

\begin{abstract}

Deep learning-based style transfer between images has recently become a popular area of research. A common way of encoding ``style'' is through a feature representation based on the Gram matrix of features extracted by some pre-trained neural network or some other form of feature statistics. Such a definition is based on an arbitrary human decision and may not best capture what a style really is. In trying to gain a better understanding of ``style'', we propose a metric learning-based method to explicitly encode the style of an artwork. In particular, our definition of style captures the differences between artists, as shown by classification performances, and such that the style representation can be interpreted, manipulated and visualized through style-conditioned image generation through a Generative Adversarial Network. We employ this method to explore the style space of anime portrait illustrations.

\end{abstract}

\input{introduction.tex}

\input{related_work.tex}

\input{metric_learning.tex}

\input{style_gan.tex}

\input{experiments.tex}

\input{result.tex}

\input{conclusion.tex}


\bibliography{anime_style_gan}
\bibliographystyle{plainnat}

\end{document}

%% file: introduction.tex
\section{Introduction}

There is an increasing interest in applying deep learning to visual arts, and
neural image style transfer techniques pioneered by Gatys and coworkers~\cite{gatys2016image} have revolutionized this area with compelling AI-driven artwork synthesis results. The simple idea of generating images by matching the convolutional features and their Gram matrices from one image to another has yield
eye-catching results not only for artistic style transfer but also for producing photorealistic results from synthetic data. Numerous follow-up work have improved the visual fidelity of the output, the speed of image generation, and the handling of multiple styles. Another line of research was image-to-image translation \cite{isola2017image}, of which 
style transfer can be though of as a special case of Generative Adversarial Networks.

While the visual quality of style transfer results keep improving, there has been
relatively few research on understanding what really is a style in the context of neural synthesis. It seems to be a consensus that ``style'' loosely equals to ``texture''.
We feel that this decision may feel a bit arbitrary, and lacking a formal understanding of the underlying mechanism. Furthermore, while in existing studies, the
representations of style is effective for neural networks, they are not intelligible to
humans.

We present a different definition and representation for artistic styles
in the context of deep learning for visual arts. We aim at a definition that is learned
from artworks and not limited to textures, as well as a representation for images that
separates style and content effectively, where the style can be interpreted
by humans. Furthermore, we train a Generative Adversarial Networks conditioned on
both style and content. Being an image generator, for which style and content can be independently controlled, it serves as a practical tool for style space visualization, as well as an alternative solution to style transfer.

Our method consists of these main steps:

\begin{itemize}
\item We train a style encoder with metric learning techniques. The goal is to encode images into a style space, such that artworks by the same artist are closer, while artworks by different artists are further apart.
\item We then train a content encoder using a Variational Autoencoder. It should in particular not encode information already described by the style encoder.
\item Finally a Generative Adversarial Network is trained conditioned on both, style and content.
\end{itemize}

To demonstrate the effectiveness our method, we explore the style space of anime portrait illustrations from a custom dataset. This level of control is not possible using existing neural synthesis approaches.

%% file: related_work.tex
\section{Related Works}
\paragraph{Style Transfer} While researches of style transfer has long existed,
of particular interest concerning the
topic of this research are the deep neural network methods based on
\cite{gatys2016image}, in which the idea
of representing style by the Gram matrices of convolutional features played a
central role.
Buiding on top of this, people have made improvements by replacing the costly optimization
process with a feedforward network (e.g. in \cite{johnson2016perceptual}), by having more
control over target location, color and scale of style \cite{gatys2017controlling}, etc.
However, these improvements did not change the central idea of representing style by the
Gram matrices.

A different approach has been to use adaptive instance normalization
\cite{dumoulin2016learned, huang2017arbitrary}. While different from the methods based
on Gram matrices, one thing they share is that the definition of style is
predetermined to be certain kinds of feature
statistics.

\paragraph{Image-to-image translation} Alternatively, style transfer can be considered
as a special case of the more general problem of image-to-image translation. First
considered for paired training images \cite{isola2017image}, the method has been developed
for unpaired training images \cite{zhu2017unpaired}. To ensure that the translation
result does have the desired style, usually a adversarial discriminator is employed to
decide if an image (is real and) has the same result as the target group of images.
Here, the definition of style is learned from the training images, but the representation
is implicit: by a discriminating process.

\paragraph{Generative Adversarial Networks and conditional GANs} Generative Adversarial
Networks (GAN) \cite{goodfellow2014generative} has been a very successful unsupervised
learning model, especially for generating realistic looking images. Numerous conditional
GAN models exist. Typically, part of the code carries some predefined meaning, and some
loss term is added to the discriminator loss that encourages the network to preserve
these information in the generated image. One example is conditioning on class label and
adding classification loss \cite{mirza2014conditional}. In our case, we condition the
GAN on style and content codes.

%% file: metric_learning.tex
\section{Metric Learning for Style}
\label{sec:metric_learning}
As discussed above, in typical neural style transfer approaches, the style is
explicitly represent by a set of numbers but the definition of style is from
an arbitrary human decision that tries to capture the texture information, while in
image-to-image type of approaches, the definition of style is learned from training
image but the representation is implicit.

We want a definition of style that is explicitly
learned to represent style, and the representation has to be a set of numbers that can
be interpreted and manipulated. Specifically, we want a style encoder which encodes images
into a style space, such that image with similar styles are encoded to points closer to
each other while images with dissimilar styles are encoded to points further away.

Such a formulation suits well in the framework of metric learning. To avoid subjective
human judgment of style, we make the assumption that artworks made by the same artist
always have similar styles while artworks made by different artists always have dissimilar
styles. This may not be exactly true, but it is a cost-effective approximation.
Now given some artworks labeled according to their author, we want to train a style
encoder network $S(\cdot)$ that minimizes

\begin{equation}
\frac{\sum_i \sum_{\mathbf{x}, \mathbf{y} \in X_i} \mathcal{L}_{\text{same}}(||S(\mathbf{x}) - S(\mathbf{y})||)}{\sum_i |X_i|^2} + \frac{\sum_{i \neq j} \sum_{\mathbf{x} \in X_i, \mathbf{y} \in X_j} \mathcal{L}_{\text{diff}}(||S(\mathbf{x}) - S(\mathbf{y})||)}{\sum_{i \neq j} |X_i| \cdot |X_j|}
\end{equation}

Where $X_i$ is the set of artworks from artist $i$, $\mathcal{L}_{\text{same}}$ and
$\mathcal{L}_{\text{diff}}$ are loss functions
(of distance) for pairs of similar styles and pairs of dissimilar styles, respectively.
We take $\mathcal{L}_{\text{same}}(t)=t^2$ and $\mathcal{L}_{\text{diff}}(t)=e^{-t^2}$.

In practice, we found that knowing only whether the two input images are of the
same style is too weak a supervision for the network. After about 50 epochs of training,
the network failed to make a significant progress. So we sought to give
it a stronger supervision.

We assume that for each artist, there is one point in style space that is
``representative'' of their style and all his artworks should be encoded to close to this
point while far from other artists. Now in addition to the style encoder $S(\cdot)$, we
learn such presumed representative styles $\mathbf{s}_i$. Together they minimize

\begin{equation}
\label{eqn:loss}
\frac{\sum_i \left(|X_i| \sum_{\mathbf{x} \in X_i} \mathcal{L}_{\text{same}}(||S(\mathbf{x}) - \mathbf{s}_i||) \right)}{\sum_i |X_i|^2} + \frac{\sum_{i \neq j} \left( |X_j|\sum_{\mathbf{x} \in X_i} \mathcal{L}_{\text{diff}}(||S(\mathbf{x}) - \mathbf{s}_j||) \right)}{\sum_{i \neq j} |X_i| \cdot |X_j|}
\end{equation}

One of our goal is to interpret the style representation. Naturally, we would
want the representation to be as simple as possible, that is to say, we want the dimension
of the style space to be small, and the dimensions should ideally be ordered by importance,
with the first dimensions accounting for style variations that most effectively
differentiate between artists. To achieve this, we use a technique called nested dropout
\cite{rippel2014learning}. The method is proposed for autoencoders but the same idea
work for discriminative problems as well. For a vector $v$, denote its projection onto the first $d$ dimensions by $v^{[d]}$. Now we define a nested dropout version of $\mathcal{L}_{\text{same}}$:

\begin{equation}
\mathcal{L}'_{\text{same}}(\mathbf{x},\mathbf{y})=\sum_{d=1}^{D}\frac{(1-t)t^{d-1}}{1-t^D}\mathcal{L}_{\text{same}}(h_d\cdot||\mathbf{x}^{[d]}-\mathbf{y}^{[d]}||)
\end{equation}

Where $h_d$ is a scale factor learned for each dimension to account for different
feature scaling under different number of dimensions, $D$ is the total number of style
dimensions, and $t$ is a hyperparameter controlling the dropout probability.
$\mathcal{L}'_{\text{diff}}$ is defined similarly with the same value for $h_d$ and $t$. In
the training, $\mathcal{L}'_{\text{same}}$ and $\mathcal{L}'_{\text{diff}}$ are used
in place of $\mathcal{L}_{\text{same}}$ and $\mathcal{L}_{\text{diff}}$ in
equation \ref{eqn:loss}. 

After training, we select an appropriate number of dimensions $D'$ for the style space such
that it is reasonably small and using only the first $D'$ dimensions performs nearly as
good as if all $D$ dimensions are used. The remaining dimensions are pruned in subsequent
steps.

%% file: style_gan.tex
\section{Style-conditioned Generative Adversarial Network}
\label{sec:style_gan}

For the second step, we want a content encoder. Variational Autoencoder
\cite{kingma2013auto} is a natural choice.
Due to the requirement that the encoder does not encode any information already encoded by
the style encoder, we made some changes: along with the output from the VAE encoder,
the output from the style encoder is provided to the decoder. In addition, similar to the
training of the style encoder, we use nested dropout: after performing reparametrization,
a prefix of random length of the output of VAE encoder is kept and the suffix is set to
all zero. Then, this is concatenated with the output from style encoder and sent to the
VAE decoder.

Let $\mathcal{L}_{\text{rec}}(\mathbf{x})$ be the reconstruction loss on input
$\mathbf{x}$, then

\begin{equation}
\mathcal{L}_{\text{rec}}(\mathbf{x}) = \sum_{d=1}^{D}\frac{(1-t)t^{d-1}}{1-t^D}||\mathrm{Dec}(S(\mathbf{x}), \mathrm{Enc}(\mathbf{x})^{[d]}) - \mathbf{x}||^2
\end{equation}

The KL-divergence part of the VAE loss remains unchanged.

Intuitively, since later dimensions has a higher chance to be dropped, the earlier dimensions
must try to learn the most import modes of variation in the image. Since the style
information is provided ``for free'', they would try to learn information not encoded
in the style. Similar to the training of style encoder above, after training the VAE, the
content encoder is pruned by testing reconstruction loss. This ensures that we only keep the
earlier dimensions that encode the content, with later dimensions that may encode redundant
information about style being discarded.

Now that we have both the style encoder and the content encoder ready, we can proceed to the
final step: a conditional Generative Adversarial Network. Let part of the input code to the
generator represent the style and let another part represent the content. In addition to
minimizing the adversarial loss, the generator tries to generate images such that on these
images the style encoder and the content encoder will give the style code and the content
code back, respectively.

The discriminator is the standard GAN discriminator:

\begin{equation}
\min_{D}\mathop{\mathbb{E}}_{\mathbf{x} \in X}[-\log D(\mathbf{x})]+\mathop{\mathbb{E}}_{\mathbf{z}\sim \mathcal{N}(0, I^{d_{\text{total}}})}[-\log (1-D(G(\mathbf{z})))]
\end{equation}

While the objective of the generator is:

\begin{align}
\min_{G}\mathop{\mathbb{E}}_{\mathbf{u}\sim \mathcal{N}(0, I^{d_{\text{style}}})}\mathop{\mathbb{E}}_{\mathbf{v}\sim \mathcal{N}(0, I^{d_{\text{content}}})}\mathop{\mathbb{E}}_{\mathbf{w}\sim \mathcal{N}(0, I^{d_{\text{noise}}})} & \left[\mathcal{L}_{\text{GAN}} + \lambda_{\text{style}}\mathcal{L}_{\text{style}} + \lambda_{\text{content}}\mathcal{L}_{\text{content}}\right]\\
\mathcal{L}_{\text{GAN}} =& -\log D(G(\mathbf{u}, \mathbf{v}, \mathbf{w}))\\
\mathcal{L}_{\text{style}} =&  ||S(G(\mathbf{u}, \mathbf{v}, \mathbf{w})) - \mathbf{u}||^2\\
\mathcal{L}_{\text{content}} =&  ||\mathrm{Enc}(G(\mathbf{u}, \mathbf{v}, \mathbf{w})) - \mathbf{v}||^2
\end{align}

where $D(\cdot)$ is the discriminator, $G(\cdot)$ is the generator, $S(\cdot)$ is the style
encoder, $\mathrm{Enc}(\cdot)$ is the content encoder (the mean of the output distribution
of VAE encoder, with variance discarded), $d_{\text{style}}$, $d_{\text{content}}$ and $d_{\text{noise}}$ are length of the parts of GAN code that is conditioned on style, on content, and unconditioned, respectively, and $\lambda_{\text{style}}$ and $\lambda_{\text{content}}$ are weighting factors. For this part, the output of $S(\cdot)$ is normalized to have zero mean and unit variance with style statistics from the training set.

%% file: experiments.tex
\section{Experiments}

We conducted our experiments on a custom dataset of anime portrait illustrations.

\subsection{Dataset}

The dataset contains about $417$ thousand anime portraits of size $256\times256$, drawn
by $2,513$ artists, obtained from the anime imageboard
Danbooru\footnote{danbooru.donmai.us}. The faces in the images are detected using AnimeFace
2009 \cite{animeface}. The faces are cropped out, rotated to upright position, padded
with black pixels if the bounding box extends beyond the border of the image, then resized
to $256\times256$. Artist information is obtained from tags on the website.
After extracting
the face patches and classifying by artist tag, we manually removed false positives of
face detection and discarded artists whose number of works is woo few (less than 50),
obtaining our dataset.

For the metric learning part, we took $10\%$ of total images or $10$ images, whichever is
larger, from each artist as the test set and use the remaining for training. For the
VAE and GAN part, we use all images for training.

\subsection{Setup}

\begin{table}
\caption{\label{tab:network} Network structure}
\begin{center}
\footnotesize
\begin{tabular}{cccc}\hline
Network        & Levels & Number of features & Number of blocks\\\hline
$S$            & $5$ & $(64, 128, 256, 384, 512)$ & $(1, 1, 1, 1, 1)$\\
$\mathrm{Enc}$, $\mathrm{Dec}$ & $6$ & $(64, 128, 192, 256, 384, 512)$ & $(1, 1, 1, 1, 1)$ \\
$G$            & $6$ & $(64, 128, 192, 256, 384, 512)$ & $(2, 2, 2, 2, 2, 2)$\\
$D$            & $4$ & $(64, 128, 256, 512)$ & $(1, 1, 1, 1)$\\
\end{tabular}
\end{center}
\end{table}

\begin{table}
\caption{\label{tab:training} Training parameters}
\begin{center}
\footnotesize
\begin{tabular}{ccccc}\hline
Training step   & Algorithm & Learning rate      & Batch size & Dropout $t$\\\hline
Style encoder   & Adam      & $10^{-4}$          & $32$       & $0.995$\\
Content encoder & RMSprop   & $5 \times 10^{-5}$ & $16$       & $0.995$\\
Conditional GAN & RMSprop   & $2 \times 10^{-5}$ & $8$        & -\\
\end{tabular}
\end{center}
\end{table}

All of our networks were based on ResNet \cite{he2016deep}.
We use weight normalized residue
blocks from \cite{xiang2017effect}. We did not use the stride-2 pooling layer after the
initial convolution layer, and our networks could have a variable number of levels
instead of a fixed $4$ levels in \cite{he2016deep}. In addition,
all networks operate in the Lab color space.

The structures of all the networks used in the experiment are listed in table
\ref{tab:network}. For VAE, in addition to the listed residue blocks, we added an
extra fully connected layer between the decoder input/encoder output and the highest
level convolutional layer, with $512$ features.

The style encoder and content encoder were both trained with $512$ output features. After
pruning, we kept the first $d_\text{style}=32$ style dimensions and the
first $d_\text{content}=128$ content dimensions
for conditional GAN training. The total number of dimensions of the GAN was also $512$,
out of which $d_\text{noise}=512-32-128=352$ dimensions were not conditioned.

The GAN discriminator operates a bit differently. We used a consortium of $3$ networks
with identical structure but operating on different scales. The networks accept
input images of size $64\times64$: the first
network sees the training images and generated images downscaled to $64\times64$; the
second network sees $4$ random $64\times64$ patches from images downscaled to
$128\times128$ and computes the average loss on the $4$ patches; the third network sees
$16$ random $64\times64$ patches from the original images and computes the average loss
on the $16$ patches. Finally, the discriminator loss is the average loss from the three
networks.

The training parameters are listed in table \ref{tab:training}. In GAN training, the
different part of the generator's loss were weighted as $\lambda_\text{style}=0.5$ and $\lambda_\text{content}=0.05$.

%% file: result.tex
\section{Results}
\label{sec:result}

\subsection{Metric Learning}
We evaluate the effectiveness of our metric learning method by considering the
classification accuracy when the style encoder is used for classification, and by 
measuring the ratio of distance between images from the same artist to the distance
between images from different artists. As a reference, we compare the results with
the same network trained on the same dataset for classification of artist.

Remember that along with the style encoder, we learned a presumed style for every
artist. So, given a style encoder trained with metric learning, we can compare
the style code of an image to the presumed styles of each artist. The image is classified
to the nearest artist.

\begin{figure}
\begin{center}
\includegraphics[width=0.45\linewidth]{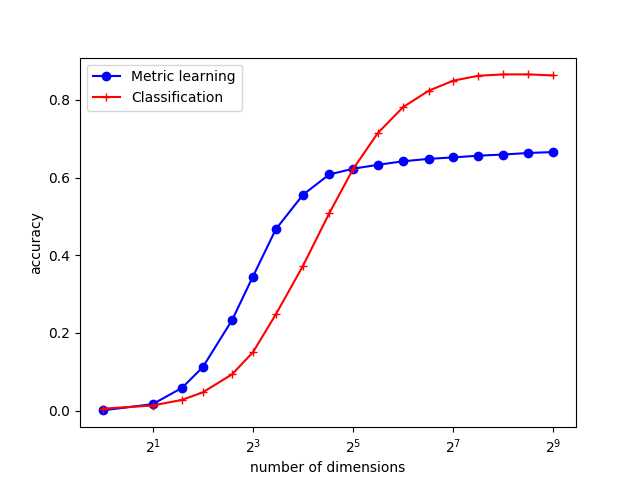}
\includegraphics[width=0.45\linewidth]{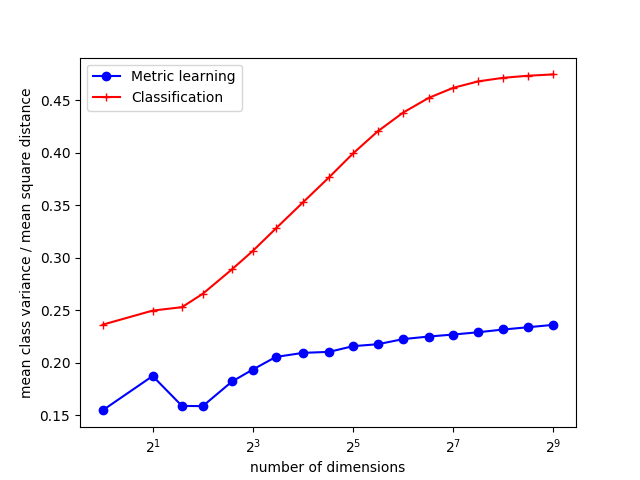}
\end{center}
\caption{\label{fig:metric_learning} Left: classification accuracy. Right: ratio of average
class variance to average squared distance between images from different classes}
\end{figure}

We trained both networks to the point when classification ceases improving. 
The left graph in figure \ref{fig:metric_learning} shows the
classification accuracy with different values of $d_\text{style}$. Note that x-axis is
in log-scale. We can see that with
a sufficient number of dimensions, the usual classification method gives better accuracy
than distance based classification on top of metric learning which is unsurprising since
in metric learning we do not directly optimize for classification accuracy. But
interestingly, when the number of dimension is very small, the metric
learning method gives better results, which shows that it uses the first dimensions
to encode style information more efficiently. We can also see that for metric
learning, using the first $32$ dimensions is almost as good as using all $512$
dimensions. Thus we decided on keeping only the first $32$ style dimensions for
subsequent steps.

As a more direct measure of whether we have achieved the goal of encoding images with
the same style to closer points in the style space, lets consider the ratio of
average variance of style of all artists to average squared distance between images
from different artists. In particular, we compute

\begin{equation}
\frac{\sum_{i}\sum_{\mathbf{x}\in X_i}||S(\mathbf{x})-\bar{\mathbf{s}}_i||^2}{\sum_{i}|X_i|}
\quad \text{and} \quad \frac{\sum_{i \neq j}\sum_{\mathbf{x}\in X_i,\mathbf{y}\in X_j}||S(\mathbf{x})-S(\mathbf{y})||^2}{\sum_{i \neq j}|X_i|\cdot|X_j|}
\end{equation}

and consider their ratio, where $X_i$ is the set of images made by artist $i$ and
$\bar{\mathbf{s}}_i=\frac{1}{|X_i|}\sum_{\mathbf{x}\in X_i}S(\mathbf{x})$ is the true
average style of artist $i$, in contrast to $\mathbf{s}_i$, the learned presumed style.
This ratio would ideally be small. The right graph in figure \ref{fig:metric_learning}
shows the ratio with different values of $d_\text{style}$. As we can see, the metric
learning method improves upon the classification method significantly, reducing the ratio by about a half.

\begin{figure}
\begin{center}
\includegraphics[width=0.45\linewidth]{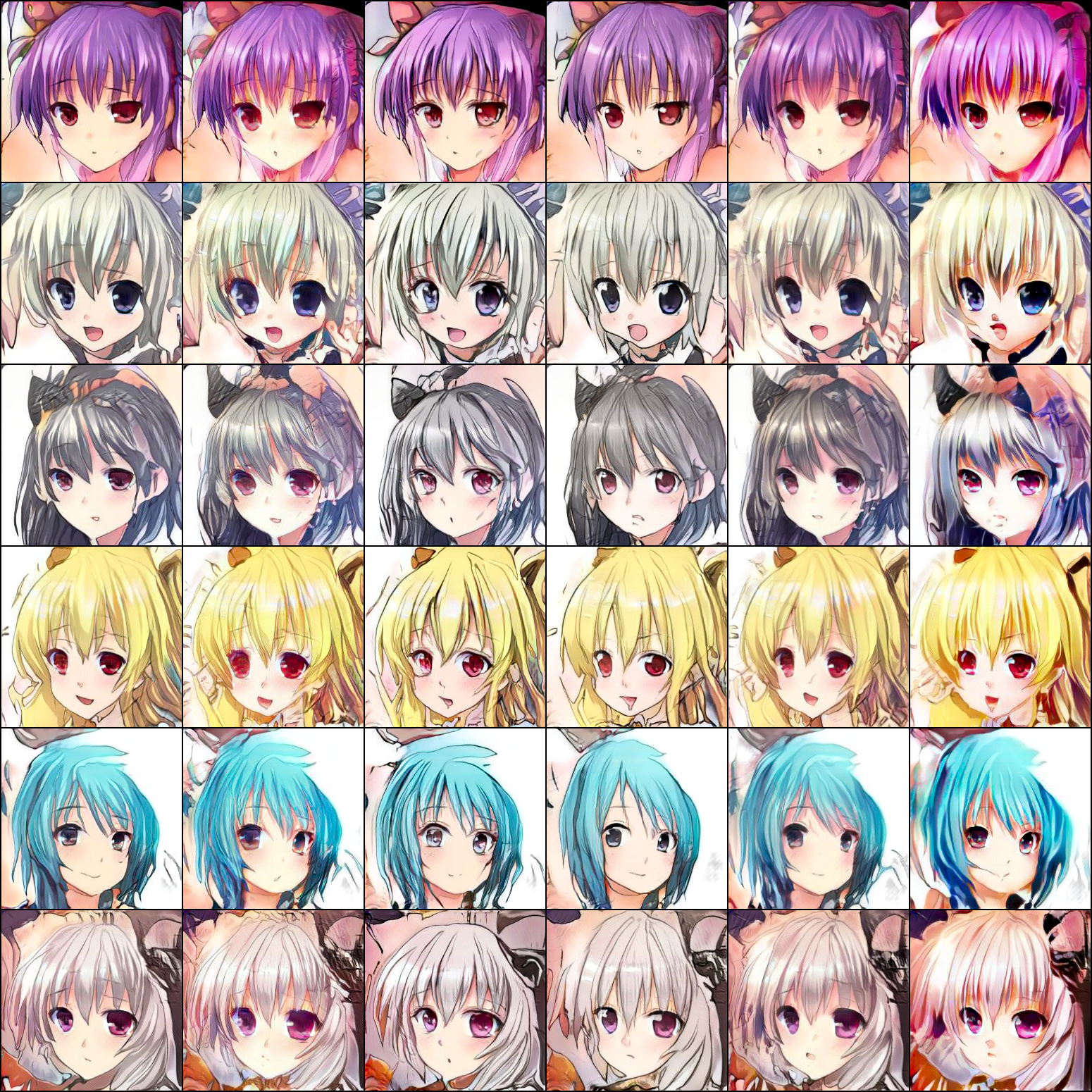}
\includegraphics[width=0.45\linewidth]{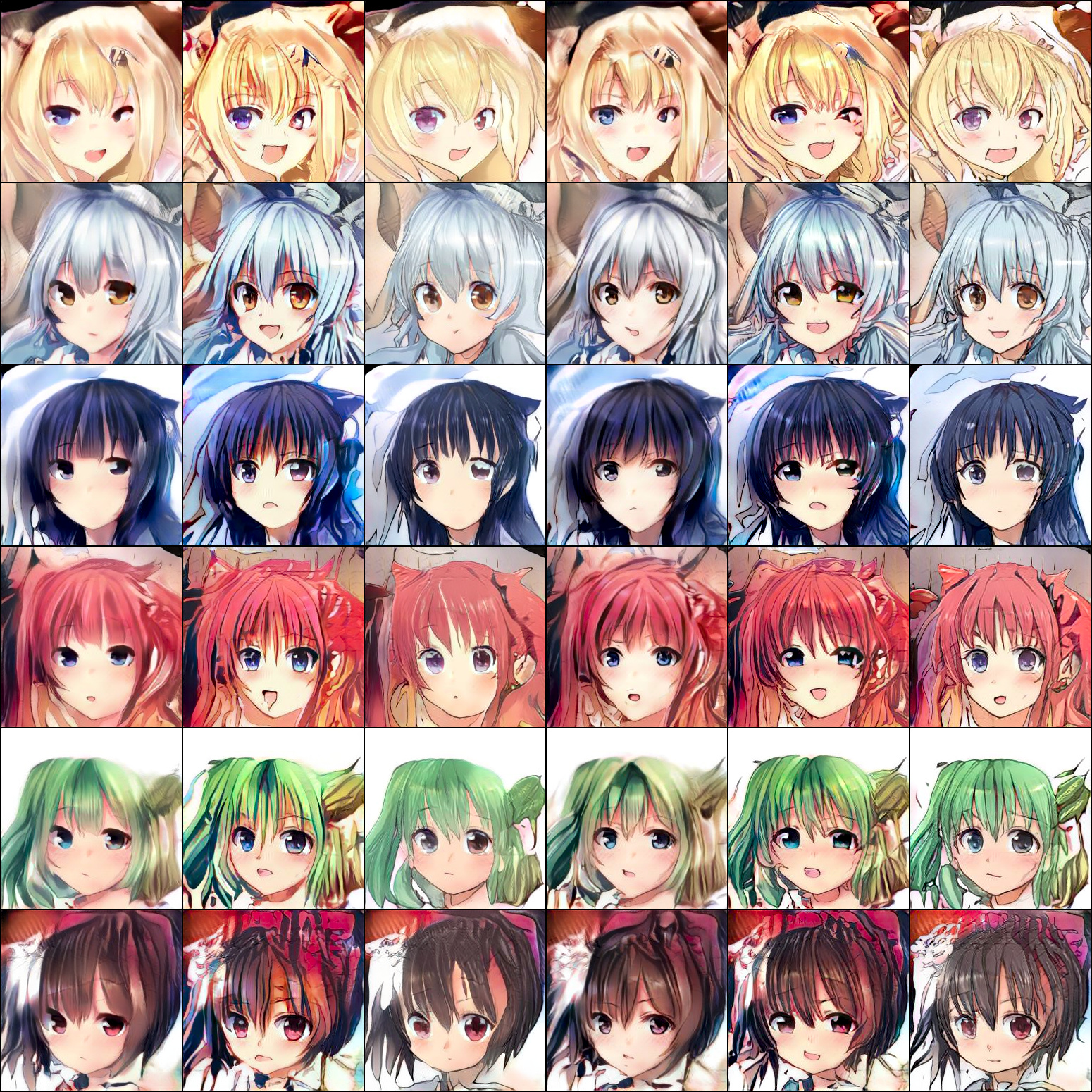}
\end{center}
\caption{\label{fig:style_vs_content} Samples generated from some combination of style codes and content codes}
\end{figure}

\subsection{Separation of Style and Content}
As a first test of our method, we would like to see whether the style dimensions and content
dimensions in the code space are actually separated. Figure \ref{fig:style_vs_content}
shows the combination of style and content. In each group, images from each row share the
content part of the code while images from each column share the style part of the code.
We can see that in each row the images depict the same character while
in each column the style of the illustration is consistent.

\subsection{Exploring the Style Space}
\begin{figure}
\begin{center}
\includegraphics[width=0.375\linewidth]{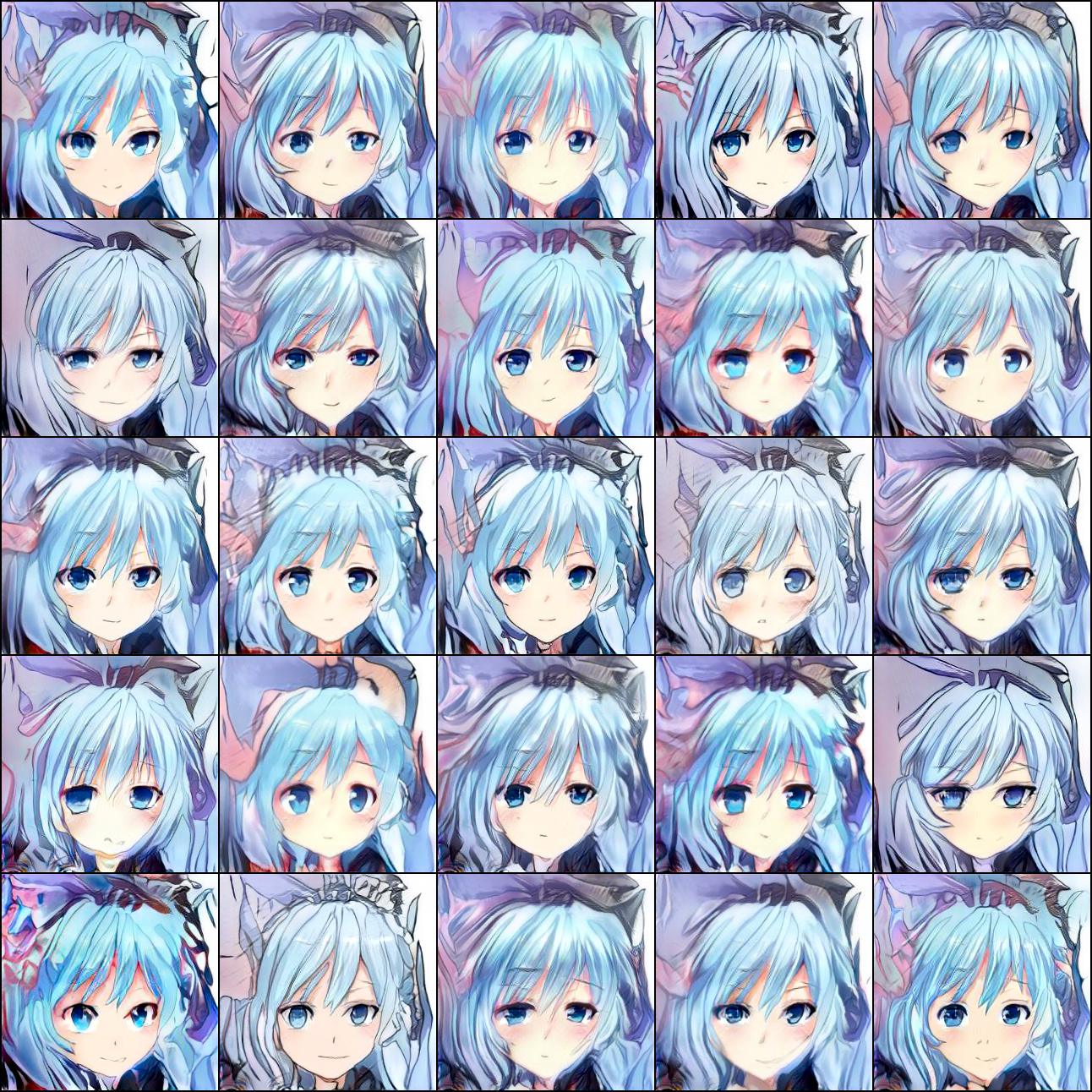}
\includegraphics[width=0.375\linewidth]{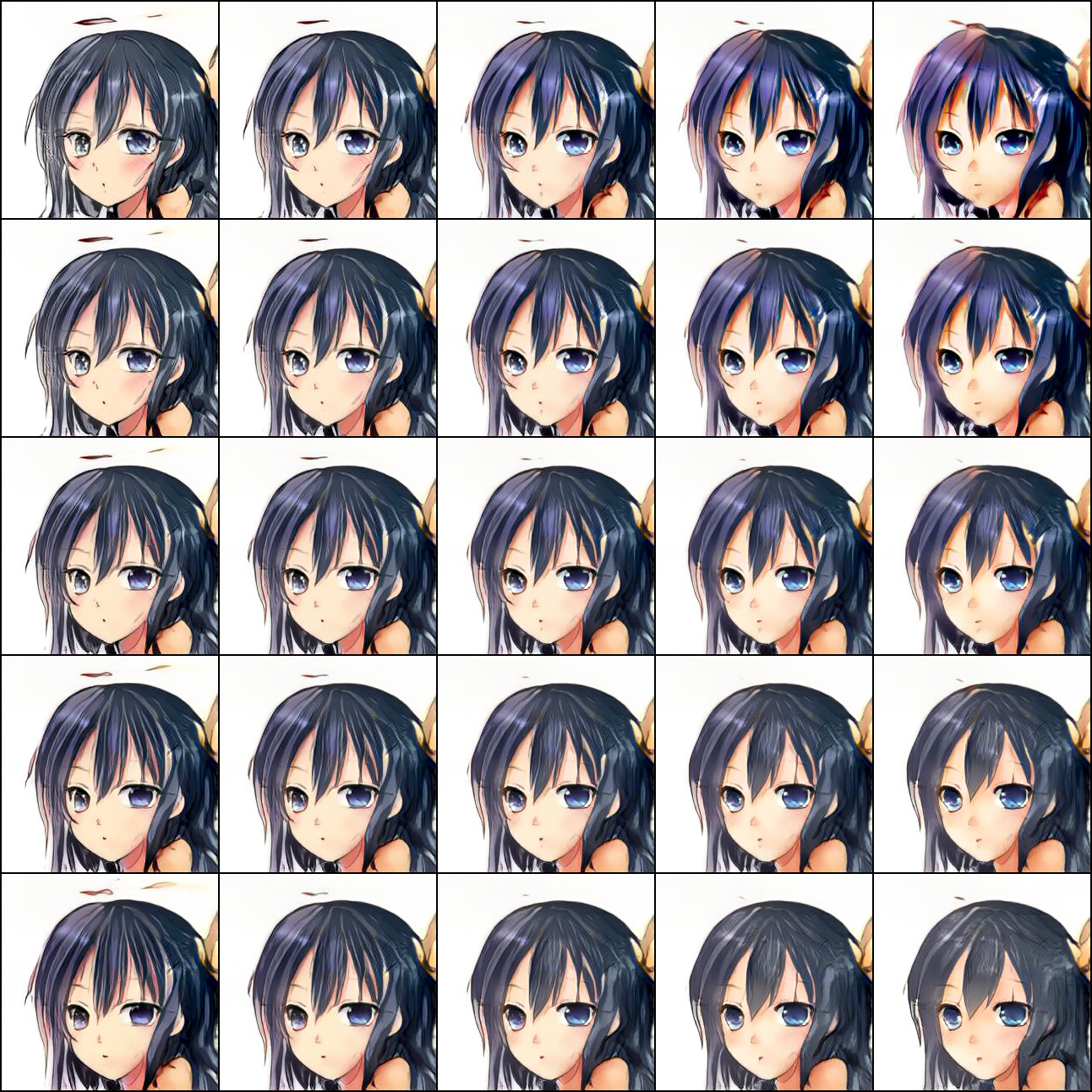}
\end{center}
\caption{\label{fig:styles} Left: random styles. Right: bilinear interpolation between styles}
\end{figure}

We show the multitude of styles that can be generated in figure \ref{fig:styles}. On the
left are samples generated from a fixed content code and random style codes. On the right,
the content code is also fixed, but the style code of the samples in the middle are
bilinear interpolated from the four images on the corner.

We would also like to know which aspects of style
are each of the dimensions controlling. For this, for each style dimension we take random
codes, fix all other dimensions and vary this style dimension and compare the generated
samples. We set the value to $-5$, $0$ and $5$. In addition, we rank all training images
along this dimension and select from lowest $10\%$ and highest $10\%$ to see if we can
observe the same variation in style.

\begin{figure}
\begin{center}
\includegraphics[width=0.675\linewidth]{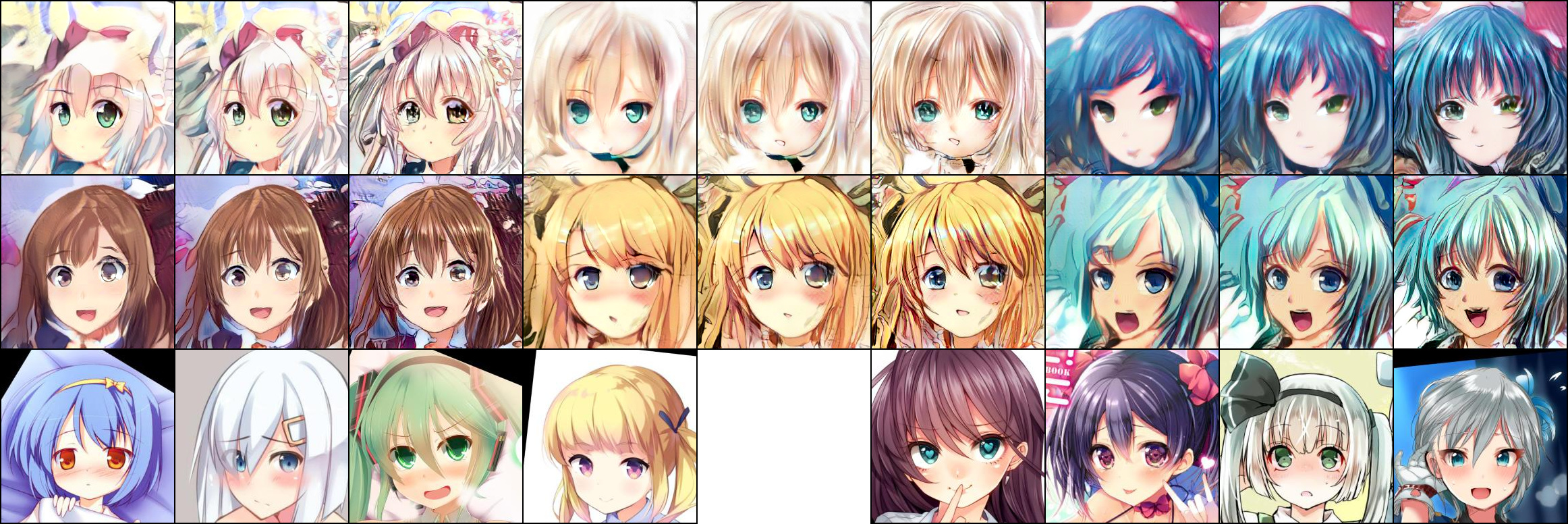}
\end{center}
\caption{\label{fig:dim_10} Effect of dimension 10}
\end{figure}

The meanings of each dimension were not as clear as we want them to be, but we were able to
explain some of them. As an example, here we show two of the
dimensions to which we can give a resonable interpretation. Figure \ref{fig:dim_10} shows the
effect of the 10th style dimension. In top two rows, the three samples in each group are
generated by setting the 10th style dimension of a random code to $-5$, $0$ and $5$ while
leaving other parts of the code unchanged. In the last row, we select images ranked lowest
and highest by 10th style dimension and show them on the left side and right side
respectively.

We found that increasing dimension 10 causes the generated samples to have finer locks of
hair and with more value contrast in the hair. Conversely, decreasing dimension 10 causes
the generated samples to have coarser locks of hair and less value contrast. The
samples from the training set agrees with this trend.

\begin{figure}
\begin{center}
\includegraphics[width=0.675\linewidth]{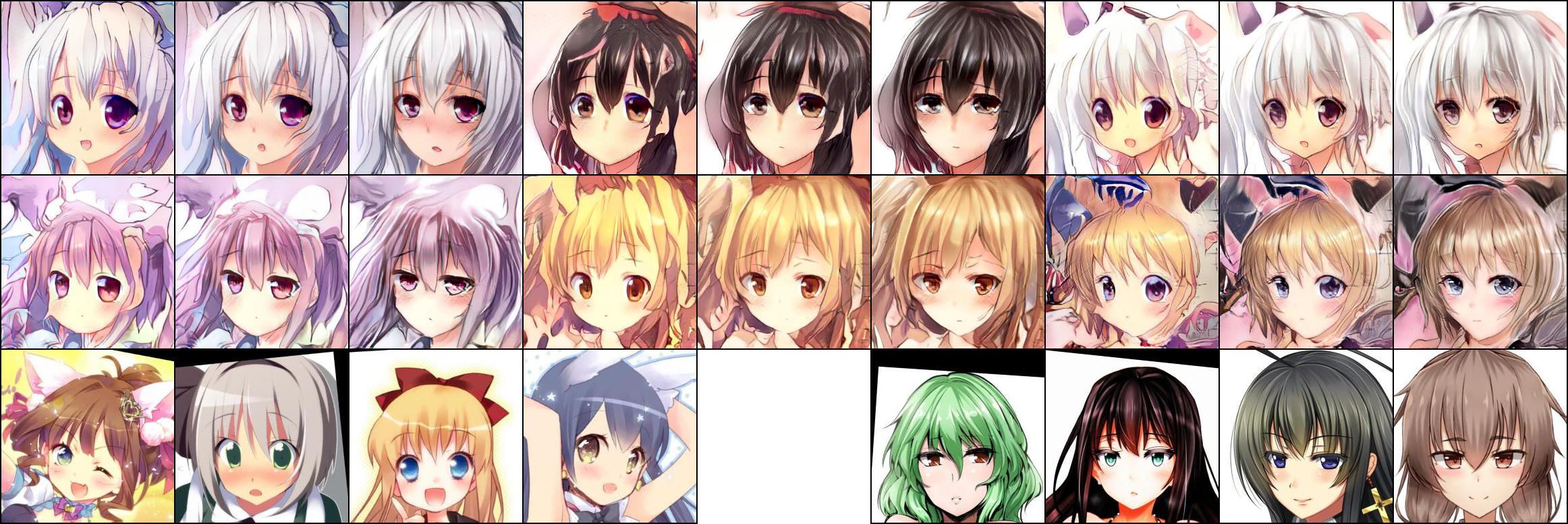}
\end{center}
\caption{\label{fig:dim_6} Effect of dimension 6}
\end{figure}

Figure \ref{fig:dim_6} shows the same experiment with the 6th style dimension. Decreasing
this dimension causes the character to look younger while increasing this dimension causes
the character to look more mature. Among other subtleties, increasing this dimension gives
the character a less round cheek, more enlongated and sharper jaw, smaller eyes and more
flatterned upper eyelids.

\subsection{Reconstruction and Style Transfer}
\begin{figure}
\begin{center}
\includegraphics[width=0.75\linewidth]{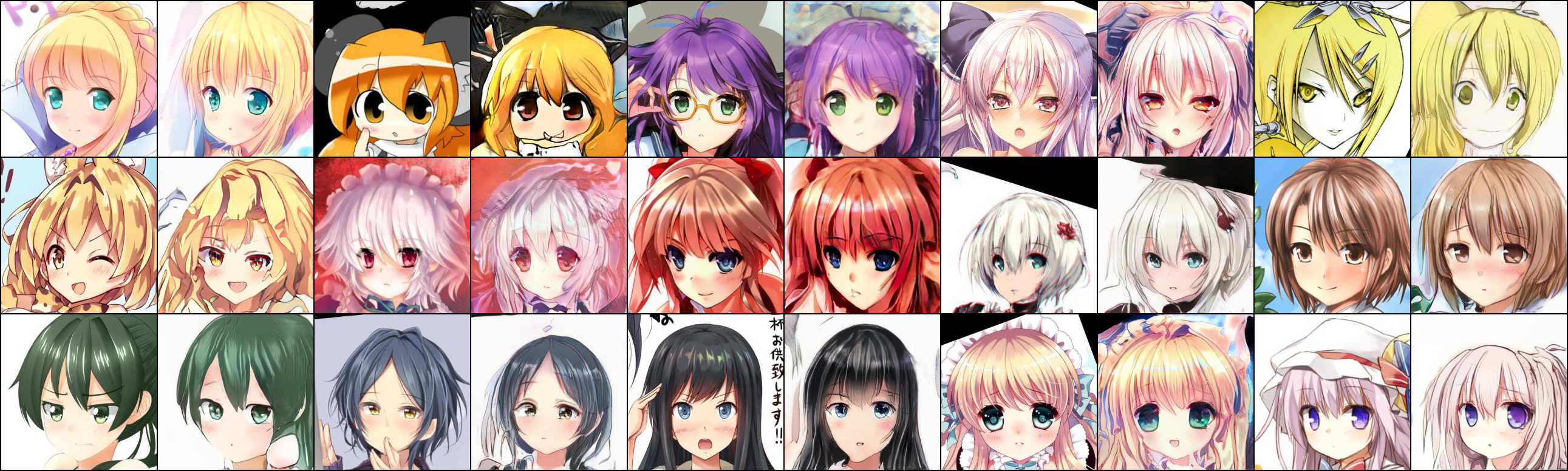}
\end{center}
\caption{\label{fig:reconstruction} Reconstruction results.}
\end{figure}

Although not trained for such a purpose, since we have a style encoder, a content encoder
and a generator conditioned on style and content, we can use these networks to reconstruct
an image by combing the output from style encoder and content encoder and sending it to
the generator, or perform style transfer between images by combining content code from
one image with style code from another image. Figure \ref{fig:reconstruction} shows
some reconstruction results. In each pair, the image on the left is from the training
images and the one on the right is the reconstruction. We can see that the network
captures lighting variations
along the horizontal direction better than along the verticle direction. In particular,
as the second pair shows, the network fails to reconstruct the horizontal stripe of
hightlight on the hair. Such styles are also noticeably absent from the random styles in
figure \ref{fig:styles}.

\begin{figure}
\begin{center}
\includegraphics[width=0.675\linewidth]{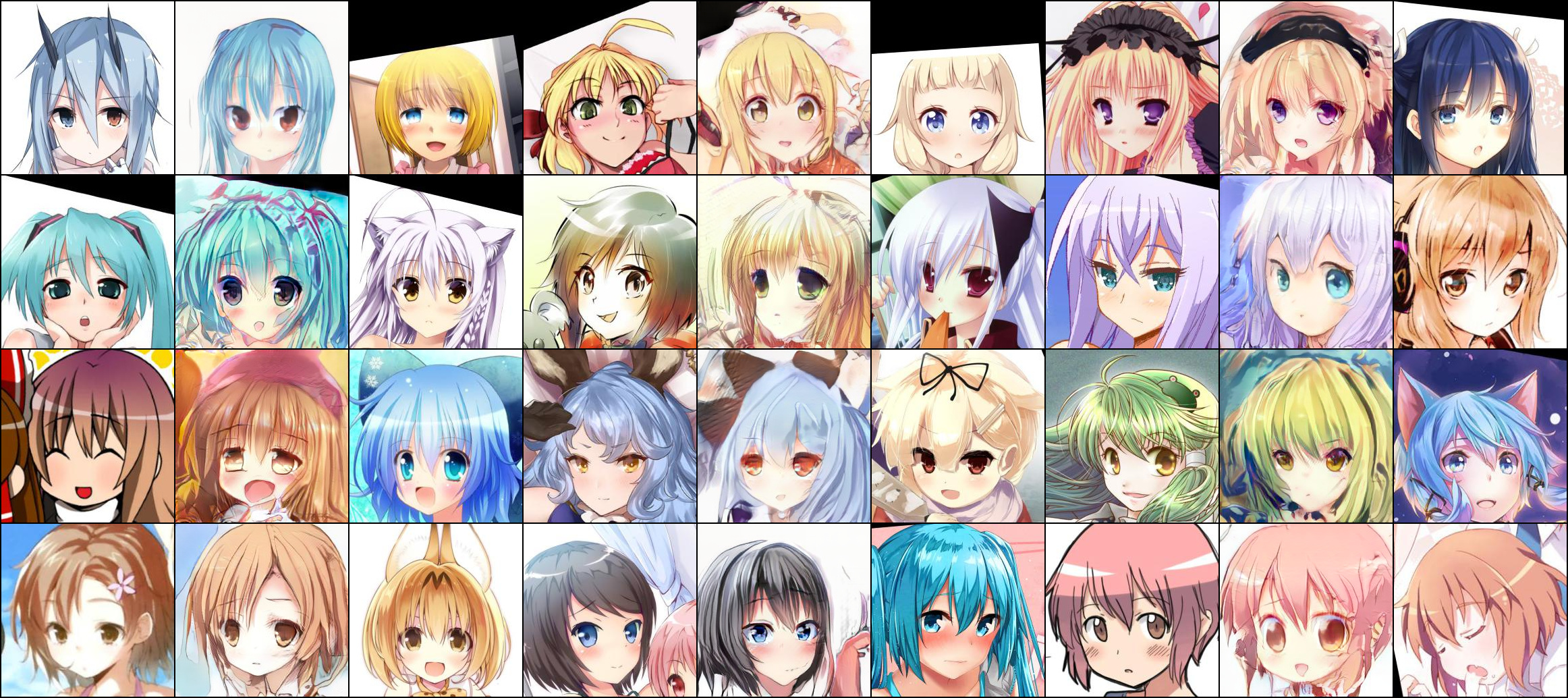}
\end{center}
\caption{\label{fig:transfer} Style transfer results.}
\end{figure}

Figure \ref{fig:transfer} shows some style transfer results. In each group of three images,
the left one and the right one are from the training set and the middle one is generated
by combining the content from the left image and the style from the right image.

%% file: conclusion.tex
\section{Conclusion}
In this paper, we presented a different view on artistic styles in deep learning. With metric
learning techniques, we obtained a style encoder that could effectively encode images into
a style space in which the distance corresponds to the similarity of style. We further
demonstrated the effectiveness of our method by visualizing and analyzing the structure of the
style space with a conditional Generative Adversarial Network. As an application of this
method, we gave an alternative solution to the problem of style transfer.

%% file: anime_style_gan.bbl
\begin{thebibliography}{14}
\providecommand{\natexlab}[1]{#1}
\providecommand{\url}[1]{\texttt{#1}}
\expandafter\ifx\csname urlstyle\endcsname\relax
  \providecommand{\doi}[1]{doi: #1}\else
  \providecommand{\doi}{doi: \begingroup \urlstyle{rm}\Url}\fi

\bibitem[Dumoulin et~al.()Dumoulin, Shlens, and Kudlur]{dumoulin2016learned}
Vincent Dumoulin, Jonathon Shlens, and Manjunath Kudlur.
\newblock A learned representation for artistic style.

\bibitem[Gatys et~al.(2016)Gatys, Ecker, and Bethge]{gatys2016image}
Leon~A Gatys, Alexander~S Ecker, and Matthias Bethge.
\newblock Image style transfer using convolutional neural networks.
\newblock In \emph{Proceedings of the IEEE Conference on Computer Vision and
  Pattern Recognition}, pages 2414--2423, 2016.

\bibitem[Gatys et~al.(2017)Gatys, Ecker, Bethge, Hertzmann, and
  Shechtman]{gatys2017controlling}
Leon~A Gatys, Alexander~S Ecker, Matthias Bethge, Aaron Hertzmann, and Eli
  Shechtman.
\newblock Controlling perceptual factors in neural style transfer.
\newblock In \emph{IEEE Conference on Computer Vision and Pattern Recognition
  (CVPR)}, 2017.

\bibitem[Goodfellow et~al.(2014)Goodfellow, Pouget-Abadie, Mirza, Xu,
  Warde-Farley, Ozair, Courville, and Bengio]{goodfellow2014generative}
Ian Goodfellow, Jean Pouget-Abadie, Mehdi Mirza, Bing Xu, David Warde-Farley,
  Sherjil Ozair, Aaron Courville, and Yoshua Bengio.
\newblock Generative adversarial nets.
\newblock In \emph{Advances in neural information processing systems}, pages
  2672--2680, 2014.

\bibitem[He et~al.(2016)He, Zhang, Ren, and Sun]{he2016deep}
Kaiming He, Xiangyu Zhang, Shaoqing Ren, and Jian Sun.
\newblock Deep residual learning for image recognition.
\newblock In \emph{Proceedings of the IEEE conference on computer vision and
  pattern recognition}, pages 770--778, 2016.

\bibitem[Huang and Belongie(2017)]{huang2017arbitrary}
Xun Huang and Serge Belongie.
\newblock Arbitrary style transfer in real-time with adaptive instance
  normalization.
\newblock In \emph{Proceedings of the IEEE Conference on Computer Vision and
  Pattern Recognition}, pages 1501--1510, 2017.

\bibitem[Isola et~al.(2017)Isola, Zhu, Zhou, and Efros]{isola2017image}
Phillip Isola, Jun-Yan Zhu, Tinghui Zhou, and Alexei~A Efros.
\newblock Image-to-image translation with conditional adversarial networks.
\newblock \emph{arXiv preprint}, 2017.

\bibitem[Johnson et~al.(2016)Johnson, Alahi, and
  Fei-Fei]{johnson2016perceptual}
Justin Johnson, Alexandre Alahi, and Li~Fei-Fei.
\newblock Perceptual losses for real-time style transfer and super-resolution.
\newblock In \emph{European Conference on Computer Vision}, pages 694--711.
  Springer, 2016.

\bibitem[Kingma and Welling(2013)]{kingma2013auto}
Diederik~P Kingma and Max Welling.
\newblock Auto-encoding variational bayes.
\newblock \emph{arXiv preprint arXiv:1312.6114}, 2013.

\bibitem[Mirza and Osindero(2014)]{mirza2014conditional}
Mehdi Mirza and Simon Osindero.
\newblock Conditional generative adversarial nets.
\newblock \emph{arXiv preprint arXiv:1411.1784}, 2014.

\bibitem[nagadomi(2017)]{animeface}
nagadomi.
\newblock Animeface 2009.
\newblock \url{https://github.com/nagadomi/animeface-2009}, 2017.
\newblock Accessed: 2018-05-15.

\bibitem[Rippel et~al.(2014)Rippel, Gelbart, and Adams]{rippel2014learning}
Oren Rippel, Michael Gelbart, and Ryan Adams.
\newblock Learning ordered representations with nested dropout.
\newblock In \emph{International Conference on Machine Learning}, pages
  1746--1754, 2014.

\bibitem[Xiang and Li(2017)]{xiang2017effect}
Sitao Xiang and Hao Li.
\newblock On the effect of batch normalization and weight normalization in
  generative adversarial networks.
\newblock \emph{arXiv preprint arXiv:1704.03971}, 2017.

\bibitem[Zhu et~al.(2017)Zhu, Park, Isola, and Efros]{zhu2017unpaired}
Jun-Yan Zhu, Taesung Park, Phillip Isola, and Alexei~A Efros.
\newblock Unpaired image-to-image translation using cycle-consistent
  adversarial networks.
\newblock In \emph{Proceedings of the IEEE Conference on Computer Vision and
  Pattern Recognition}, pages 2223--2232, 2017.

\end{thebibliography}
